\documentclass[letterpaper]{article} 
\usepackage{aaai24}
\usepackage{times}  
\usepackage{helvet}  
\usepackage{courier}  
\usepackage[hyphens]{url}  
\usepackage{graphicx} 
\usepackage{natbib}  
\usepackage{caption} 
\usepackage{subcaption}
\usepackage{multirow}
\newcommand\Tstrut{\rule{0pt}{2ex}}       

\usepackage{algorithm}
\usepackage{algpseudocode}

\usepackage{amsmath,amssymb}

\usepackage{tikz}
\usetikzlibrary{positioning}
\usetikzlibrary{3d} 
\usetikzlibrary{decorations.pathreplacing,calligraphy}
\usetikzlibrary{arrows}
\usetikzlibrary{shapes}

\tikzstyle{sum}=[draw, rounded corners, very thick, rectangle, minimum height=10em, minimum width=5.1em, fill=gray!20]
\tikzstyle{transformer}=[draw, rounded corners= 1.5em, very thick, rectangle, minimum height=22em, minimum width=12em, fill=gray!20]
\tikzstyle{MLP}=[draw, rounded corners, very thick, rectangle, minimum height=2em, minimum width=8em, fill=blue!20]
\tikzstyle{norm}=[draw, rounded corners, very thick, rectangle, minimum height=2em, minimum width=8em, fill=green!20]
\tikzstyle{head}=[draw, rounded corners, very thick, rectangle, minimum height=4em, minimum width=8em, fill=orange!20]
\tikzstyle{el}=[draw, very thick, ellipse , minimum height=38em, minimum width=10em, fill opacity=0.2]
\tikzstyle{connection}=[ultra thick,every node/.style={sloped,allow upside down},draw=\edgecolor,opacity=0.7]

\usepackage{pgfplots}
\pgfplotsset{width=10cm,compat=1.9}

\usepackage{newfloat}
\usepackage{listings}
\DeclareCaptionStyle{ruled}{labelfont=normalfont,labelsep=colon,strut=off} 
\floatstyle{ruled}
\newfloat{listing}{tb}{lst}{}
\floatname{listing}{Listing}
\title{Blacksmith: Fast Adversarial Training of Vision Transformers via a Mixture of Single-step and Multi-step Methods}

\author{
    Mahdi Salmani\equalcontrib,
    Alireza Dehghanpour Farashah\equalcontrib,
    Mohammad Azizmalayeri,
    Mahdi Amiri,
    Navid Eslami,
    Mohammad Taghi Manzuri,
    Mohammad Hossein Rohban
}
\affiliations{

    \{mahdi.salmani, alireza.dehghanpour, m.azizmalayeri, mahdiamiri, navid.eslami, manzuri, rohban\}@sharif.edu
}

\usepackage{bibentry}

\begin{document}

\maketitle

\begin{abstract}
Despite the remarkable success achieved by deep learning algorithms in various domains, such as computer vision, they remain vulnerable to adversarial perturbations. Adversarial Training (AT) stands out as one of the most effective solutions to address this issue; however, single-step AT can lead to Catastrophic Overfitting (CO). This scenario occurs when the adversarially trained network suddenly loses robustness against multi-step attacks like Projected Gradient Descent (PGD). 
Although several approaches have been proposed to address this problem in Convolutional Neural Networks (CNNs), we found out that they do not perform well when applied to Vision Transformers (ViTs).
In this paper, we propose \textit{Blacksmith}, a novel training strategy to overcome the CO problem, specifically in ViTs. Our approach utilizes either of PGD-2 or Fast Gradient Sign Method (FGSM) randomly in a mini-batch during the adversarial training of the neural network. This will increase the diversity of our training attacks, which could potentially mitigate the CO issue. To manage the increased training time resulting from this combination, we craft the PGD-2 attack based on only the first half of the layers, while FGSM is applied end-to-end. Through our experiments, we demonstrate that our novel method effectively prevents CO, achieves PGD-2 level performance, and outperforms other existing techniques  including N-FGSM, which is the state-of-the-art method in fast training for CNNs.
\end{abstract}

\section{Introduction}

\begin{figure*}
    \centering
    \includegraphics[page=1,width=\textwidth]{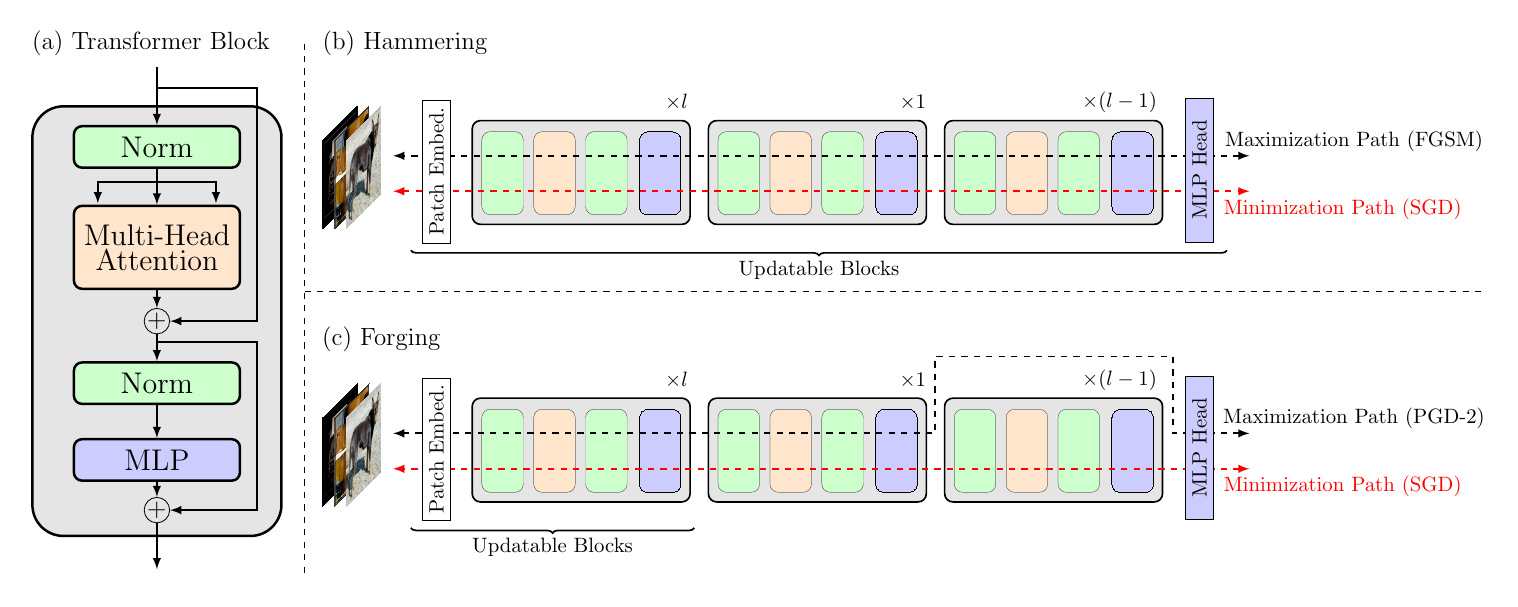}
    \caption{(a) Structure of a standard Transformer Block. (b) Hammering Scheme: Focuses on the training of the last layers of the transformer while maintaining the robustness of the first layers. A single-step method is appliend on all layers, mainly FGSM or RS-FGSM, and thus takes as much time as a normal FGSM. (b) Forging Scheme: Focuses on the first layers of the model, increasing their robustness. A multi-step method, namely PGD-2, is applied on the first $l+1$ layers, bypassing the last $l-1$ layers straight into the head of the network, in order to find an adversarial sample. The resulting sample is then forwarded through the network, and the gradient is used to update the weights of only the first $l$ layers. Note that the head of the entire network only uses the class token embedding, which is derived from the embeddings of the previous layer. Therefore, $l+1$ layers was used to calculate the attack sample, so that all weights in the first $l$ layers affect the gradient. This entire process is almost as fast as a normal application of FGSM.}
    \label{fig:key_idea}
\end{figure*}

Deep Neural Networks (DNNs) have enjoyed remarkable success in various domains, such as Computer Vision \cite{vision1, vit}, Natural Language Processing \cite{transformer}, Speech Processing \cite{speech}, etc. This success has been further amplified by the advent of the transformer architecture, which gave birth to Vision Transformers (ViTs) \cite{vit}. ViTs employ a self-attention mechanism on sequences of image patches in order to solve machine learning problems better than the familiar Convolutional Neural Networks (CNNs) \cite{ResNet}. 

Unfortunately, DNNs remain susceptible to adversarial perturbations \cite{perturbation_cite}, indicating a lack of robustness when the test samples are slightly perturbed. To address this vulnerability, many solutions have been explored by researchers, with Adversarial Training (AT) \cite{AT} and Certified Robustness \cite{certified1, certified2, certified3} being the most prominent. The main goal in AT is to augment the training samples with  adversarial perturbations, while the certified robustness techniques come up with a perturbation radius for any given test sample that the model output is guaranteed to remain unchanged for the perturbed input \cite{fan2021adversarial}. 

The issue with adversarial training is that it involves obtaining adversarial perturbations for each input on every epoch. 
This slows down the training process due to the iterative nature of methods that are used to obtain the adversarial samples, such as the Projected Gradient Descent (PGD)~\cite{kurakin2018adversarial, AT}. PGD typically involves numerous ($\sim$10) backpropagation steps to craft the adversarial perturbation in a gradient ascent process. Although many efforts have been made to improve the quality of AT \cite{zhang2020attacks, shaeiri, Wang2020Improving, hat}, leading to stronger defences, the adversarial training long running time has remained less explored. 

Most efforts along fast adversarial training have been mainly focused  on variations of single-step PGD attacks, such as Fast Gradient Sign Method (FGSM) \cite{fgsm}, to achieve robustness while reducing the training time. However, \cite{CO} showed that training with single-step attack methods have a failure mode called catastrophic overfitting (CO), where the model abruptly loses its robustness to multi-step attacks early in the training process. This limitation prevents us from using these efficient approaches for achieving robustness against strong adversarial attacks. 

Although several approaches have been proposed to address the CO issue in fast AT of CNNs \cite{free,  fast, gradalign, ZeroGrad, NFGSM}, we found that they do not perform well when applied to ViTs. We discover that all of these methods suffer from the CO problem in ViTs. We even observe the CO problem on a network adversarially trained with PGD-2 attack in some occasions. 
It is worth mentioning that multi-step AT can be employed with ViTs as well as the CNNs \cite{shao2022on, herrmann2022pyramid,VIT_adv1}, yet the challenge of the CO issue for single-step attacks still persists within ViTs. This raises an important issue in fast adversarial training of ViTs, which is further emphasized by their superior performance when compared to CNNs.

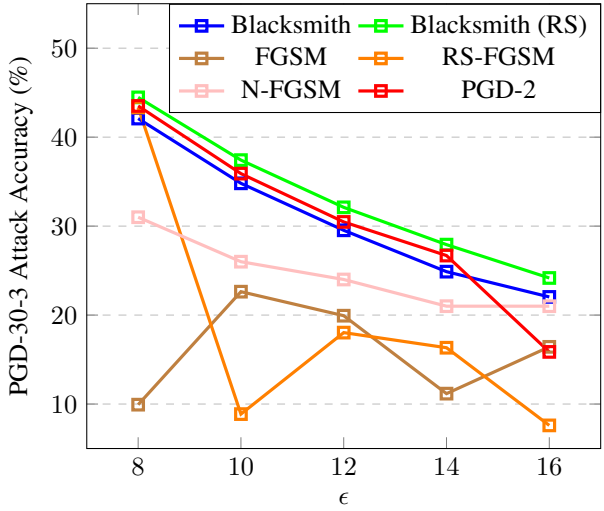
\begin{figure}[ht]
    \centering
    \begin{tikzpicture}[scale=1.0]
        \begin{axis}[
            height=7.5cm,
            width=\columnwidth,
            xlabel={$\epsilon$},
            ylabel={PGD-30-3 Attack Accuracy (\%)},
            xmin=7, xmax=17,
            ymin=5, ymax=55,
            xtick={8, 10, 12, 14, 16},
            ytick={10, 20, 30, 40, 50},
            legend style={legend columns=2, at={(0.16,0.877)}, anchor=west},
            ymajorgrids=true,
            grid style=dashed,
            every axis plot/.append style={very thick}
        ]
        
        \addplot[
            color=blue,
            mark=square
            ]
            coordinates {
            (8,42.09)
            (10,34.81)
            (12,29.53)
            (14,24.88)
            (16,22.04)
            };

        \addplot[
            color=green,
            mark=square
            ]
            coordinates {
            (8,44.47)
            (10,37.42)
            (12,32.11)
            (14,27.92)
            (16,24.17)
            };

        \addplot[
            color=brown,
            mark=square
            ]
            coordinates {
            (8,9.93)
            (10,22.63)
            (12,19.93)
            (14,11.17)
            (16,16.43)
            };

        \addplot[
            color=orange,
            mark=square
            ]
            coordinates {
            (8,43.33)
            (10,8.87)
            (12,18.03)
            (14,16.33)
            (16,7.6)
            };

        \addplot[
            color=pink,
            mark=square
            ]
            coordinates {
            (8,31)
            (10,26)
            (12,24)
            (14,21)
            (16,21)
            };

        \addplot[
            color=red,
            mark=square
            ]
            coordinates {
            (8,43.51)
            (10,35.9)
            (12,30.47)
            (14,26.69)
            (16,15.86)
            };
            
            \legend{Blacksmith, Blacksmith (RS), FGSM, RS-FGSM, N-FGSM, PGD-2}
        \end{axis}
    \end{tikzpicture}
    \caption{CIFAR10 classification accuracy of the baseline methods as a function of perturbation size ($\epsilon$). Blacksmith (RS) refers to the instance of Blacksmith which uses RS-FGSM in the Hammering scheme instead of FGSM. As the figure shows, Blacksmith achieves the best overall results.}
    \label{fig:pgd_accuracies}
\end{figure}

To this end, we propose \textit{Blacksmith}, a novel fast training procedure that prevents the CO problem. Blacksmith operates on ViTs, exploiting the assumption that the inputs and outputs of every layer have the same shape, a characteristic seen in ViTs, but not in CNNs. An overall view of our proposed method is depicted in Fig. \ref{fig:key_idea}. Our proposed method consists of two main operations. In the \textit{Hammering} operation, the ViT is adversarially trained end-to-end using FGSM, whereas in the \textit{Forging} operation, PGD-2 is used to train the first half of the network. At each step of the training, one of these operations is selected randomly. Our main insight is that one should give higher priority to the first half layers of the network in making the model robust. This is evident as it becomes increasingly challenging for the second half layers of the network to recover a perturbed sample if the first half layers have struggled in doing so. Therefore, the stronger and more time consuming PGD-2 attack is applied to the first half layers to prioritize them more. This scheme causes the model to become robust to multi-step attacks, while retaining the speed of fast adversarial training without being catasrophically overfit.

Our contributions can be summarized as follows:
\begin{itemize}

\item We evaluate state-of-the-art methods for fast training of CNNs on ViTs. We observe that all of these methods suffer from the CO problem, which introduce a barrier for fast adversarial training of ViTs.

\item We propose Blacksmith, a novel fast adversarial training algorithm suitable for ViTs, achieving PGD-2 level performance with FGSM level training time. Moreover, we provide insights into the design decisions behind Blacksmith and the robustness that it brings.

\item We rigorously evaluate our method on the CIFAR-10 and CIFAR-100 datasets using PGD-30-3 and Auto-Attack \cite{autoattack}. Our results show that our proposed method performs well on ViTs, achieving up to $1.5 \times$ better accuracy relative to N-FGSM \cite{NFGSM}, which is the state-of-the-art in fast adversarial training of CNNs. 

\end{itemize}

All experiments were conducted on NVIDIA Tesla P100 GPUs. Fig.~\ref{fig:pgd_accuracies} gives an overview of the adversarial performance of Blacksmith, alongside the baseline methods.

\section{Related Work}
 To overcome the CO failure mode, several modifications to the original FGSM method have been proposed \cite{free}. FGSM-RS \cite{fast} proposed to use random starts when performing FGSM on a given training input, while ZeroGrad \cite{ZeroGrad} stabilizes the training process by zeroing out gradient dimensions with small absolute value when obtaining perturbations during training. The intuition here is based on the fact that small gradient dimensions are more likely to change their signs in consecutive epochs, which results in large weight updates. GradAlign \cite{gradalign} claims that the loss gradient with respect to the input varies significantly around a clean sample, resulting in unreliable FGSM attacks. Therefore, they used a regularization term to smoothen the loss function so that a linear form would better approximate the loss, which in turn allows  single-step attacks to find better perturbations. N-FGSM \cite{NFGSM} added larger initial noises compared to the FGSM-RS. They showed that larger input noise implicitly makes the loss landscape linear, and hence makes loss linearization in the FGSM attack work. They also did not clip the perturbation when it exceeds the given perturbation norm bound. The rationale here is that using clipping, the contribution of the FGSM perturbation compared to the initial random noise reduces when the noise magnitude grows. This leads to the attack behaving like a random perturbation for large initial noises, and hence decreasing its effectiveness.

\section{Preliminaries}
Despite the great success of DNNs in various domains, they are not robust to adversarial perturbations. AT \cite{AT} is one of the most successful methods to overcome this problem. Assume that $f_\theta(x)$ represents a neural network with the input $x$, and parameters $\theta$. With this notation, we can define the adversarial training problem as
\begin{equation}
   \min_{\mathbf \theta} \mathbb{E}_{{\mathbf x}, y \sim \mathcal{D}} (\max_{{\|\delta\|}_\infty \leq \epsilon} \mathcal{L} (f_{\mathbf \theta}({\mathbf x}+\delta), y)),
\end{equation}
where $\delta$ is a perturbation with $\ell_\infty$-norm less than $\epsilon$, $y$ is the label, and $\mathcal{L}$ denotes the cross-entropy loss function, which is a common choice for the classification tasks. The aim of the inner maximization is to find the worst-case perturbation, maximizing the loss function, while the outer minimization tries to update the weight to minimize it. This min-max optimization encourages model to learn features that are robust even against the worst-case perturbations \cite{ilyas2019adversarial, tao2021better}.

The minimization phase is applied as in the case of standard training. The inner maximization problem, however, can be solved via iterative gradient-guided mothods. PGD is among the most successful methods in solving this maximization problem. Its update rule is defined as 
\begin{equation}
   \delta = \delta + \alpha \nabla_x (\mathcal{L} (f_{\mathbf \theta}(x +\delta), y)),
\end{equation}
where $\alpha$ is the step size in updating the attack perturbation. Note that the initial value of $\delta$ is determined by the algorithm, with many papers using random initializations at the beginning of the procedure. Due to the iterative nature of the PGD algorithm, the adversarial training time will increase linearly w.r.t the number of iterations. To combat this drawback, FGSM applies a single step attack with the update rule of 
\begin{equation}
   \delta = \delta_{init} + \alpha \cdot sign[\nabla_x (\mathcal{L} (f_{\mathbf \theta}(x +\delta), y))],
\end{equation}
where $\delta_{init}$ is the initial value of $\delta$ which depends on the algorithm, and $sign(\cdot)$ is the point-wise sign function that returns 1 for positive numbers, and -1 otherwise. Note that here the perturbation size $\epsilon$ is equal to the value of step-size $\alpha$. The problem with FGSM and its variants is that the network trained with fast methods could suddenly loose their robustness to multi-step attacks \cite{CO}. This creates a challenge for us in adversarially training DNNs using the fast training methods.  

\section{Proposed Method}
Our own experience indicates that single step methods tend to be unstable in many situations, leading to catastrophic overfitting of the models (as seen in Fig.~\ref{fig:pgd_accuracies}). Motivated by this observation, we mix multi-step approaches with fast methods, achieving more stable fast solutions. The key idea is to make the first layers of the transformer more robust via a multi-step approach, while training the last layers using a fast method. This, coupled with the randomization of the aforementioned schemes, cuts down on the running time of multi-step approaches, while maintaining their performance. An overview of the aforementioned points is presented in Fig.~\ref{fig:key_idea}. This paper's insights, the aforementioned schemes, and the algorithm as a whole are presented in the following sections.

\subsection{Intuition}
In the discussion that follows, we denote the ViT model being studied as $f_{1:2l}$ with patch embedding $PE$, transformer layers $L_1, \dots, L_{2l}$, and head $H$. To gain an intuition about the ``overall robustness-degree" of layer $k$, we define
\begin{equation}
    \gamma_k = ||v_k^{Attack} - v_k^{Original}||_p,
\end{equation}
where $v_i^{Original}$ is the output of $L_i$ when $f_{1:2l}$ is fed a random batch $\mathcal{B}$ from the training dataset, and $v_i^{Attack}$ is the output of $L_i$ when fed the PGD-30 sample batch $\mathcal{B}$. In our setting, $p$ is chosen to be $2$. Using this notation, we define
\begin{equation}
    \gamma^* = \frac{\gamma_{2l}}{\gamma_l}.
\end{equation}
One can argue that $\gamma^*$ gives a sense of how much the last $l$ layers of $f_{1:2l}$ ``amplify" the perturbations (resulting from an adversarial attack on the network) in the output of the first $l$ layers, which are then carried over to the end of the transformer layers. That is to say, if $\gamma^*$ is small, one can conjecture that $L_{l+1}, \dots, L_{2l}$ have become robust w.r.t the output space of $L_l$, as they control the extent of the attack. 

\begin{figure}[t]
    \centering
    \begin{tikzpicture}
        \begin{axis}[
            height=6cm,
            width=\columnwidth,
            xlabel={Epoch No.},
            ylabel={$\gamma^*$},
            xmin=0, xmax=31,
            ymin=0, ymax=18,
            xtick={0, 5, 10, 15, 20, 25, 30},
            ytick={2, 4, 6, 8, 10, 12, 14, 16},
            legend pos=north east,
            ymajorgrids=true,
            grid style=dashed,
            every axis plot/.append style={very thick}
        ]
        \addplot[
            color=pink,
            mark=square
        ]
        coordinates {
        (1, 2.230850603668456)
        (2, 2.0618330682752495)
        (3, 1.9085002182247697)
        (4, 2.2892140143673467)
        (5, 3.722197453692015)
        (6, 5.538027217396436)
        (7, 4.718530226729266)
        (8, 6.000706679715831)
        (9, 13.86153814512629)
        (10, 12.558793163863008)
        (11, 11.08640799184613)
        (12, 9.007368477647528)
        (13, 10.560208208063731)
        (14, 14.098729137184812)
        (15, 16.758162766137403)
        (16, 14.23181162442686)
        (17, 15.665779141020357)
        (18, 11.418179353671457)
        (19, 10.778907762371611)
        (20, 9.937155709911062)
        (21, 8.965312333689882)
        (22, 11.210689458088503)
        (23, 6.4615269037169005)
        (24, 6.801504838130793)
        (25, 7.484851254140935)
        (26, 6.930644909758797)
        (27, 5.585671356391806)
        (28, 5.390650697782995)
        (29, 4.816525416219198)
        (30, 5.934159959539004)
        };

            \addplot[
            color=red,
            mark=square
            ]
            coordinates {
            (1,1.873145574087705)
            (2,2.797823113185146)
            (3,2.3891950601985554)
            (4,8.795390201928901)
            (5,9.483399014365906)
            (6,10.988811367677746)
            (7,12.227812970242741)
            (8,12.32030916592477)
            (9,12.979950508024327)
            (10,11.51426303992519)
            (11,12.320831907303555)
            (12,13.184301454505645)
            (13,10.453533469493163)
            (14,10.645439739940265)
            (15,10.004257806639123)
            (16,9.140668134915252)
            (17,9.324024297305005)
            (18,7.890052800255966)
            (19,7.504640996561731)
            (20,6.741921050462661)
            (21,6.890064609415435)
            (22,7.004205407258858)
            (23,6.728327502170163)
            (24,6.795771233463909)
            (25,6.219782115270181)
            (26,7.074805650422644)
            (27,7.1235450711998896)
            (28,6.780459347930524)
            (29,6.672994369047594)
            (30,7.161803716950611)
            };

            \addplot[
            color=black,
            mark=square
            ]
            coordinates {
            (1,1.8982435985615638)
            (2,1.8065619173178598)
            (3,2.0713628347994284)
            (4,2.617342322502721)
            (5,3.802479741972305)
            (6,5.49758823205091)
            (7,7.453816900934985)
            (8,8.250616703015185)
            (9,8.570680814920966)
            (10,8.147165215040294)
            (11,8.654077547670669)
            (12,8.747072475899333)
            (13,7.092293430500227)
            (14,7.2633693511587705)
            (15,6.996275649258292)
            (16,6.161764732537755)
            (17,6.502216626431796)
            (18,4.978608506749978)
            (19,5.203706074617374)
            (20,4.633156329990934)
            (21,4.341397050471921)
            (22,5.687051691484271)
            (23,4.829893206454155)
            (24,4.320942418805481)
            (25,4.342972567393502)
            (26,4.701967070789783)
            (27,4.781785704239618)
            (28,4.99306517527003)
            (29,4.731912618887013)
            (30,5.28693980969131)
            };

             \legend{N-FGSM, PGD-2, PGD-10}
        \end{axis}
    \end{tikzpicture}
    \caption{The ratio $\gamma^*$ while training on CIFAR10 with perturbation size 8, as training epochs go by. Lower values of $\gamma^*$ indicate that an algorithm has made the last $l$ layers of the transformer more robust w.r.t the output of the first $l$ layers. PGD-10 exhibits lower values of $\gamma^*$ compared to PGD-2 and N-FGSM.
    }
    \label{fig:L2_ratios}
\end{figure}
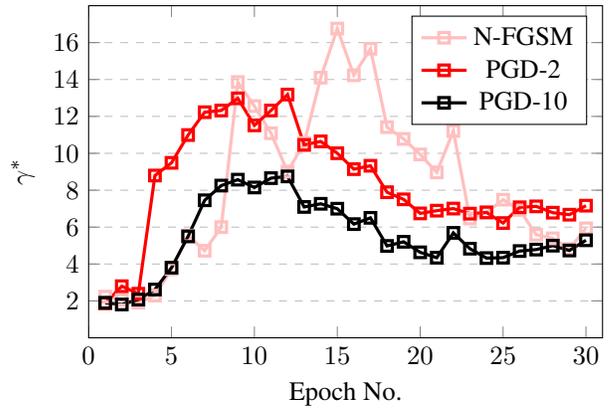

Fig.~\ref{fig:L2_ratios} shows $\gamma^*$ for N-FGSM, PGD-2, and PGD-10. Clearly, increasing the number of the training attack steps has reduced $\gamma^*$ during the training, which aligns well with the fact that models trained with higher attack step numbers are more robust in general. 
Moreover, the value of $\gamma^*$ suddenly increases in the case of N-FGSM, maintaining its overall magnitude. This happens at the 8th epoch, which is also the same epoch that the CO problem is encountered in.

Based on these intuitions and observation, since there is no general consensus about the cause of the CO problem in the scientific community, we opt to make our algorithm behave similar to iterative methods (such as PGD), which are known to experience CO less frequently. Using the Forging scheme, Blacksmith is made similar to PGD-2 in the sense that it can search through the attack sample-space using an iterative scheme while preserving the computational cost.

It should be noted that the definition of $\gamma^*$ can be extended to calculate a measure of robustness for every individual layer of the network. We avoid considering these ratios, however, to keep our study as simple as possible. 

In essence, Blacksmith manages to control the values of $\gamma^*$ and $\gamma_l$ by the use of its two schemes. The Forging scheme controls the value of $\gamma_l$, since it directly applies PGD-2 to $L_1, \dots, L_l$, applying more attack iterations and making the network behave similarly to the PGD case. Furthermore, Hammering works towards a low $\gamma^*$, which correlates to keeping $\gamma_{2l}$ close to $\gamma_l$ when coupled with the Forging scheme.

\subsection{Hammering}
The first scheme used in Blacksmith is Hammering (Fig.~\ref{fig:key_idea}-(b)). Hammering focuses on the last layers of a transformer, making them robust while preserving the robustness of the first layers of the network. It starts by doing a complete forward pass of a minibatch through the network. It then applies a backward step on the network to find adversarial examples to train all of the model layers with. The Hammering algorithm is outlined in Algorithm~\ref{alg:hammering}.

\begin{algorithm}[t]
\caption{Hammering algorithm.}
\label{alg:hammering}
\begin{algorithmic}[1]
\State \textbf{Given:} Minibatch ($\mathcal{B}$), Clip ($c \in \{T, F\}$), Model ($f_{1:2l}$)
\State $g \gets \nabla_{\mathcal{B}} \mathcal{L}(f_{1:2l}(\mathcal{B} + \mathbf{Unif}(-k\epsilon, k\epsilon)))$
\State $\mathcal{B'} \gets \mathcal{B}$
\If{$c$} \Comment{Single-Step Sampling Algorithm}
    \State $\mathcal{B'} \gets \Pi_{Ball(\mathcal{B}, \epsilon)} (\mathcal{B} + \alpha \cdot sign(g))$
\Else
    \State $\mathcal{B'} \gets \mathcal{B} + \alpha \cdot sign(g)$
\EndIf
\State Update $PE, L_1, \dots, L_{2l}, H$ with $\mathcal{B'}$
\end{algorithmic}
\end{algorithm}

One can see that Hammering is a generalization of the known single-step methods. For example, setting $k=2$ and $c=F$ results in N-FGSM. Because of this, it exhibits the same execution time.

\subsection{Forging}
The second scheme used in Blacksmith is Forging. The key idea is to use a multi-step approach to make the first layers of the transformer robust. I.e., keep $\gamma_l$ from rapidly increasing.

A general overview of Forging is presented in Fig.~\ref{fig:key_idea}-(c). The procedure starts by doing a forward pass through $L_1, \dots, L_{l+1}$. The result of these layers is then forwarded straight to the head of the network, disregarding the remaining layers, leading to a final prediction. Then, the input gradient is calculated via back-propagation through only these $l+1$ layers. This procedure is repeated two times, updating the current samples in order to reach a final set of adversarial inputs.

Since the last $l-1$ layers are completely dismissed in both the forward and backward pass, the entire procedure is two times faster than a full PGD-2, and as fast as a full single-step method. The resulting adversarial samples are used to train the first $l$ layers of the network. That is, the last $l$ layers of the network are freezed. This effectively applies a two-step method on the first layers of the network. The Forging algorithm is presented in Algorithm~\ref{alg:forging}.

Note that the head of a vision transformer only operates on the class token \cite{vit}. Therefore, if the class token of the $k$-th layer is forwarded to the head of the network, the network paths of the first $k$ layers will not be fully considered in creating the resulting perturbed samples. That is, only the computations that lead to the creation of the class token will be considered. Knowing this, the result of the first $l+1$ layers is chosen to be forwarded, since the full effect of the first $l$ layers is desired.

\begin{algorithm}[t]
\caption{Forging algorithm.}
\label{alg:forging}
\begin{algorithmic}[1]
\State \textbf{Given:} Minibatch ($\mathcal{B}$), Clip ($c \in \{T, F\}$), Model ($f_{1:2l}$) 
\State $\mathcal{B'} \gets \mathcal{B} + \mathbf{Unif}(-k\epsilon, k\epsilon)$
\For{$i=1 \dots 2$} \Comment{Multi-Step Sampling Algorithm}
    \State $g \gets \nabla_{\mathcal{B}} \mathcal{L}(f_{1:l+1}(\mathcal{B'}))$
    \If{$c$}
        \State $\mathcal{B'} \gets \Pi_{Ball(\mathcal{B'}, \epsilon)} (\mathcal{B'} + \frac{\alpha}{2} \cdot sign(g))$
    \Else
        \State $\mathcal{B'} \gets \mathcal{B'} + \frac{\alpha}{2} \cdot sign(g)$
    \EndIf
\EndFor
\State Update $PE, L_1, \dots, L_{l}$ with $\mathcal{B'}$
\end{algorithmic}
\end{algorithm}

\subsection{Blacksmith}

\begin{algorithm}[t]
\caption{Blacksmith training algorithm.}
\label{alg:blacksmith}
\begin{algorithmic}[1]
\State \textbf{Given:} Forge-Rate ($\lambda$), Epoch Count ($N$), Dataset ($\mathcal{D}$)
\For{$i=1 \dots N$}
    \ForAll{minibatch $\mathcal{B} \subset \mathcal{D}$}
        \State $p \sim \mathbf{Unif}(0, 1)$
        \If{$p < \lambda$}
            \State Apply \emph{Forging} with $\mathcal{B}$ \Comment{Algorithm~\ref{alg:forging}}
        \Else
            \State Apply \emph{Hammering} with $\mathcal{B}$ \Comment{Algorithm~\ref{alg:hammering}}
        \EndIf
    \EndFor
    \State $\lambda \gets f_\lambda(\lambda)$ \Comment{Forge-Rate Scheduler}
\EndFor
\end{algorithmic}
\end{algorithm}

The Blacksmith training algorithm is outlined in Algorithm~\ref{alg:blacksmith}. It utilizes a Forge-Rate Scheduler, which changes the value of $\lambda$ as the epochs complete. This value is then used to interleave Forging and Hammering, making the different layers of the model robust. 

\section{Experiments}
We choose FGSM, RS-FGSM, N-FGSM, and PGD-2 as the baseline attacks to compare Blacksmith with. We considered GradAlign to be in our baseline attacks as well, but decided against it due to the fact that 1) it is slow, taking more time than PGD-2, 2) it is a regularizer-based method that can be applied in the context of any attack.

Furthermore, two different versions of Blacksmith are considered for experimentation, namely, Blacksmith and Blacksmith (RS). The former chooses its parameters such that its Hammering stage behaves as FGSM would, while the latter chooses its parameters to use RS-FGSM.

In the next subsections, our experimental setup is presented, followed by a discussion of the results achieved, including adversarial accuracies and running times.

\begin{table*}[t]
    \centering
    \renewcommand{\arraystretch}{1.2}
    \caption{Clean and adversarial evaluation results on the CIFAR10 dataset, while varying the perturbation size ($\epsilon$). The best adversarial accuracies are shown in boldface font, which are all achieved by our method.}
    \label{tab:CIFAR10_results}
    \begin{tabular}{c|c||@{\hskip 0.125cm}c@{\hskip 0.25cm}c@{\hskip 0.25cm}c@{\hskip 0.25cm}c@{\hskip 0.25cm}c@{\hskip 0.125cm}}
        Method & Evaluation Mode & $\epsilon = 8$ & $\epsilon = 10$ & $\epsilon = 12$ & $\epsilon = 14$ & $\epsilon = 16$ \\
        \hline \\[-0.41cm] \hline
        \multirow{2}{*}{FGSM} & Clean (\%) & $39.37 \pm 15.51$ & $56.0 \pm 28.78$ & $61.43 \pm 22.31$ & $32.03 \pm 12.81$ & $33.13 \pm 8.0$ \Tstrut \\
         & Adversarial (\%) & $9.93 \pm 14.05$ & $22.63 \pm 16.08$ & $19.93 \pm 14.1$ & $11.17 \pm 8.66$ & $16.43 \pm 2.22$ \Tstrut \\
         \hline
         \multirow{2}{*}{RS-FGSM} & Clean (\%) & $88.7 \pm 0.62$ & $49.73 \pm 7.76$ & $46.83 \pm 27.98$ & $56.27 \pm 12$ & $43.03 \pm 14.2$ \Tstrut \\
         & Adversarial (\%) & $43.23 \pm 0.93$ & $8.87 \pm 12.54$ & $18.03 \pm 13.17$ & $16.33 \pm 11.58$ & $7.6 \pm 10.75$  \Tstrut \\
         \hline
         \multirow{2}{*}{N-FGSM} & Clean (\%) & $67.93 \pm 23.83$ & $63.73 \pm 15.98$ & $42.98 \pm 10.6$ & $38.61 \pm 19.41$ & $36.99 \pm 3.49$ \Tstrut \\
         & Adversarial (\%) & $31.4 \pm 22.2$ & $26.27 \pm 18.58$ & $24.19 \pm 5.03$ & $20.84 \pm 6.14$ & $21.28 \pm 1.18$ \Tstrut \\
         \hline
         \multirow{2}{*}{\textbf{Blacksmith}} & Clean (\%) & $88.7 \pm 0.12$ & $86.08 \pm 0.16$ & $82.55 \pm 0.16$ & $77.99 \pm 0.69$ & $72.49 \pm 1.03$ \Tstrut \\
         & Adversarial (\%) & $42.09 \pm 0.6$ & $34.81 \pm 0.27$ & $29.53 \pm 0.53$ & $24.88 \pm 0.29$ & $22.04 \pm 0.43$ \Tstrut \\
         \hline
         \multirow{2}{*}{\textbf{Blacksmith (RS)}} & Clean (\%) & $87.38 \pm 0.19$ & $83.95 \pm 0.2$ & $79.19 \pm 0.33$ & $72.6 \pm 1.55$ & $66.89 \pm 1.01$ \Tstrut \\
         & Adversarial (\%) & $\mathbf{44.47 \pm 0.31}$ & $\mathbf{37.42 \pm 0.41}$ & $\mathbf{32.11 \pm 0.09}$ & $\mathbf{27.92 \pm 0.52}$ & $\mathbf{24.17 \pm 0.52}$ \Tstrut \\
         \hline \\[-0.41cm] \hline
         \multirow{2}{*}{PGD-2} & Clean (\%) & $89.99 \pm 0.08$ & $87.8 \pm 0.05$ & $84.31 \pm 0.13$ & $79.47 \pm 0.56$ & $64.56 \pm 12.22$ \Tstrut \\
         & Adversarial (\%) & $43.51 \pm 0.13$ & $35.9 \pm 0.15$ & $30.47 \pm 0.15$ & $26.69 \pm 0.42$ & $15.86 \pm 11.17$ \Tstrut \\
    \end{tabular}
\end{table*}

\subsection{Setup}
\begin{table*}[t]
    \centering
    \renewcommand{\arraystretch}{1.2}
    \caption{Clean and adversarial evaluation results on the CIFAR100 dataset, while varying the perturbation size ($\epsilon$). The best adversarial accuracies are shown in boldface font, which are all achieved by our method.}
    \label{tab:CIFAR100_results}
    \begin{tabular}{c|c||@{\hskip 0.125cm}c@{\hskip 0.25cm}c@{\hskip 0.25cm}c@{\hskip 0.25cm}c@{\hskip 0.25cm}c@{\hskip 0.125cm}}
        Method & Evaluation Mode & $\epsilon = 8$ & $\epsilon = 10$ & $\epsilon = 12$ & $\epsilon = 14$ & $\epsilon = 16$ \\
        \hline \\[-0.41cm] \hline
        \multirow{2}{*}{FGSM} & Clean (\%) & $49.97 \pm 22.53$ & $24.9 \pm 5.87$ & $43.53 \pm 14.51$ & $40.6 \pm 17.25$ & $15.9 \pm 3.05$ \Tstrut \\
         & Adversarial (\%) & $16.43 \pm 11.62$ & $0.0 \pm 0.0$ & $10.47 \pm 7.42$ & $9.0 \pm 6.36$ & $0.0 \pm 0.0$ \Tstrut \\
         \hline
         \multirow{2}{*}{RS-FGSM} & Clean (\%) & $66.5 \pm 1.34$ & $56.47 \pm 6.14$ & $58.87 \pm 9.18$ & $63.67 \pm 2.38$ & $63.3 \pm 0.73$ \Tstrut \\
         & Adversarial (\%) & $8.83 \pm 10.52$ & $0.03 \pm 0.05$ & $0.0 \pm 0.0$ & $0.0 \pm 0.0$ & $0.0 \pm 0.0$ \Tstrut \\
         \hline
         \multirow{2}{*}{N-FGSM} & Clean (\%) & $54.08 \pm 10.69$ & $48.02 \pm 13.92$ & $42.57 \pm 11.18$ & $37.29 \pm 16.51$ & $39.92 \pm 9.96$ \Tstrut \\
         & Adversarial (\%) & $15.64 \pm 11.08$ & $12.14 \pm 9.18$ & $6.38 \pm 8.68$ & $10.29 \pm 7.28$ & $11.69 \pm 1.78$ \Tstrut \\
         \hline
         \multirow{2}{*}{\textbf{Blacksmith}} & Clean (\%) & $68.32 \pm 0.29$ & $64.8 \pm 0.09$ & $60.78 \pm 0.29$ & $56.74 \pm 0.22$ & $52.34 \pm 0.39$ \Tstrut \\
         & Adversarial (\%) & $22.29 \pm 0.36$ & $18.71 \pm 0.27$ & $15.42 \pm 0.05$ & $13.04 \pm 0.31$ & $11.21 \pm 0.2$ \Tstrut \\
         \hline
         \multirow{2}{*}{\textbf{Blacksmith (RS)}} & Clean (\%) & $66.52 \pm 0.31$ & $62.18 \pm 0.07$ & $57.86 \pm 0.22$ & $53.31 \pm 0.52$ & $48.11 \pm 1.1$ \Tstrut \\
         & Adversarial (\%) & $\mathbf{24.58 \pm 0.16}$ & $\mathbf{20.06 \pm 0.23}$ & $\mathbf{16.96 \pm 0.28}$ & $\mathbf{14.51 \pm 0.12}$ & $\mathbf{12.52 \pm 0.15}$ \Tstrut \\
         \hline \\[-0.41cm] \hline
         \multirow{2}{*}{PGD-2} & Clean (\%) & $68.68 \pm 0.16$ & $65.54 \pm 0.09$ & $62.71 \pm 0.12$ & $59.66 \pm 0.07$ & $55.84 \pm 0.41$ \Tstrut \\
         & Adversarial (\%) & $23.39 \pm 0.14$ & $18.73 \pm 0.05$ & $15.5 \pm 0.11$ & $13.35 \pm 0.08$ & $11.66 \pm 0.1$  \Tstrut \\
    \end{tabular}
\end{table*}

\begin{figure}[t]
    \centering
    \begin{tikzpicture}[scale=1.0]
        \begin{axis}[
            height=6cm,
            width=\columnwidth,
            xlabel={Epoch No.},
            ylabel={Learning Rate},
            xmin=0, xmax=30,
            ymin=0, ymax=0.25,
            xtick={0, 5, 10, 15, 20, 25, 30},
            ytick={0, 0.05, 0.1, 0.15, 0.2},
            yticklabel style={/pgf/number format/fixed,
                  /pgf/number format/precision=3},
            legend pos=north east,
            ymajorgrids=true,
            grid style=dashed,
            every axis plot/.append style={very thick}
        ]
        
        \addplot[
            color=blue,
            ]
            coordinates {
            (0,0)
            (7.5,0.1)
            (15,0.2)
            (22.5,0.1)
            (30,0)
            };

        \addplot[
            color=orange,
            dash pattern={on 7pt off 7pt on 7pt off 7pt}
            ]
            coordinates {
            (0,0)
            (7.5,0.2)
            (15,0.2)
            (22.5,0.1)
            (30,0)
            };
            
            \legend{Hammering, Forging}
        \end{axis}
    \end{tikzpicture}
    \caption{Learning rate values of Hammering and Forging, as training epochs go by. Hammering uses a cyclic learning rate scheduler, while Forging uses a trapezoidal scheduler, which reaches higher values more quickly.}
    \label{fig:lr_scheduler}
\end{figure}
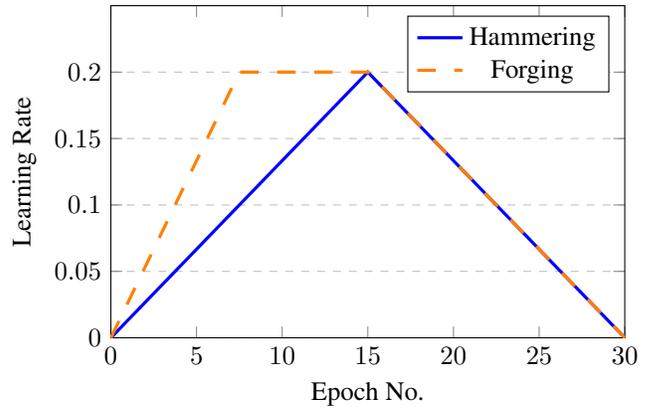

To compare Blacksmith with other single-step methods, two training alternatives exist: 1) using a conventional cyclic learning rate scheduler as in \cite{NFGSM}, or 2) using a multi-step learning rate scheduler as in \cite{VIT_adv2}. The best of these strategies is chosen for every method for comparison's sake. In other words, FGSM, RS-FGSM and N-FGSM use a multi-step scheduler, while PGD-2 and both Blacksmiths use a cyclic scheduler. Moreover, for the best performance, Blacksmith and Blacksmith (RS) use step-sizes $\alpha=\epsilon$ and $\alpha=1.25\epsilon$, respectively, where $\epsilon$ is the perturbation size of the method. More details can be found in the Appendix.

This paper assumes a separate optimizer for each of the Forging and Hammering schemes. The rationale behind this decision is to separate the momentum calculations for the schemes, which could otherwise interfere with each other. The Forging scheme uses a cyclic learning rate scheduler, with a maximum learning rate of $0.2$, while the Hammering scheme uses a trapezoidal scheduler with the same maximum learning rate. The latter scheduler brings up the learning rate more quickly than the former, preventing catastrophic overfitting. They both, however, reach a shared learning rate in the long run, as shown in Fig. \ref{fig:lr_scheduler}.

The Forge-Rate considered should be large enough so that the first transformer layers achieve an initial degree of robustness. This is imperative, because the model has a high chance of diverging otherwise. It should be gradually lowered, however, to raise the sample efficiency of the training and increasing the clean accuracy. To adhere to these constraints, our experiments first have $\lambda=0.66$ for half of the total epochs. It will then be lowered to $\lambda=0.33$ for the rest of the training, increasing the frequency of Hammerings. For a more detailed discussion on the choice of $\lambda$, see Appendix.

All of the methods considered are evaluated on both CIFAR10 and CIFAR100 \cite{cifar} while varying the value of perturbation size ($\epsilon$), once using a standard procedure and once under a PGD-30-3 attack, finetuning ViT-B with 12 layers ($l=6$) for 30 epochs. Each measurement is taken from 3 separate experiments with different seeds, all on NVIDIA Tesla P100s.

Note that the experiments with perturbation size $\epsilon$ actually use step-size $\alpha=\frac{\epsilon}{2}$ for PGD-2. This is because PGD-2 performs poorly while using $\alpha=\epsilon$, and thus, replacing it with $\frac{\epsilon}{2}$ makes for a better baseline (See Appendix).

\subsection{Results}

Tables~\ref{tab:CIFAR10_results} and \ref{tab:CIFAR100_results} present the experiment results of our baselines, as well as two versions of Blacksmith, on the CIFAR10 and CIFAR100 datasets, respectively. As shown in these tables, both instances of Blacksmith outperform all other single-step methods by a large margin, achieving over four times better accuracy in some cases. They manage to achieve up to $ 1.5 \times$ better results when compared with N-FGSM, the current state-of-the-art single-step adversarial training algorithm, on both datasets considered. Furthermore, Blacksmith and Blacksmith (RS)'s adversarial accuracy is on-par with PGD-2, with the latter dominating it in all settings.

Although some single-step methods achieve comparable results to Blacksmith when using $\epsilon=8$ in Table~\ref{tab:CIFAR10_results}, they fail when larger perturbation sizes are involved. Under these conditions, the CO problem becomes commonplace among FGSM and its variants, making their results inferior to that seen from Blacksmith.

Since classifying CIFAR100 images is inherently harder than CIFAR10, both clean and adversarial accuracies reported in Table~\ref{tab:CIFAR100_results} are lower than their CIFAR10 counterparts. The added difficulty also means that the CO problem occurs more often, even causing $0.0\%$ adversarial accuracy across all experiment seeds for methods like FGSM and RS-FGSM.


In contrast, both Blacksmith methods considered exhibit a very small standard deviation when compared to all other methods mentioned, which indicates that it is less susceptible to the catastrophic overfitting problem, which can even happen to PDG-2 when the perturbation size is largest. In fact, the proposed methods did not catastrophically overfit in any of the experiments conducted, while all other methods experienced this at least once.

It should be noted that although Blacksmith (RS) has better adversarial performance when compared with the base Blacksmith, it does suffer from a slightly lower clean accuracy. This signifies the trade-off between clean and adversarial accuracy in these two methods \cite{tsipras2018robustness, trades}.

To show that Blacksmith is robust against other attacks as well, it is evaluated using Auto-Attack \cite{autoattack}. The results of these experiments are provided in Table~\ref{tab:autoattack}. This table shows that Blacksmith holds its own against Auto-Attack, maintaining high adversarial accuracies and not suffering from the CO problem. Since the other baseline methods experienced CO in this setting, there is no mention of them in Table~\ref{tab:autoattack}.

\begin{table}[t]
    \centering
    \renewcommand{\arraystretch}{1.2}
    \caption{Vanilla Blacksmith's robustness when evaluated with Auto-Attack, as training conditions change. The measurements provided are calculated across three separate experiment with different seeds.}
    \label{tab:autoattack}
    \begin{tabular}{c||c@{\hskip 0.5cm}c}
         Dataset & $\epsilon=8$ & $\epsilon=16$ \\
         \hline
         CIFAR10 & $40.50 \pm 0.67\%$ & $16.75 \pm 0.59\%$ \Tstrut \\
         CIFAR100 & $21.03 \pm 0.51\%$ & $8.25 \pm 0.48\%$ \Tstrut \\
    \end{tabular}
\end{table}

\subsection{Computational Complexity}
Table~\ref{tab:training_times} presents the total training times of Blacksmith alongside our baselines. The measurements show that Blacksmith beats PGD-2's running time by a large margin, and even slightly outperforms other single-step methods. This runtime reduction from the other single-step methods is caused by freezing $L_{l+1}, \dots, L_{2l}$ in the Forging scheme, which bypasses parts of the gradient computation in PyTorch. This is in contrast with the other single-step methods, which have to do these computations regardless. Therefore, one can see that Blacksmith brings adversarial accuracies on par with PGD-2 to the table, without incurring much overhead.

\begin{table}[t]
    \centering
    \caption{Total training time measurements, when training on the CIFAR10 dataset.}
    \label{tab:training_times}
    \begin{tabular}{c|c}
         Method & Total Training Time ($s$) \\
         \hline
         FGSM & $14315$ \Tstrut \\
         RS-FGSM & $14327$ \Tstrut \\
         N-FGSM & $14282$ \Tstrut \\
         PGD-2 & $19921$ \Tstrut \\
         PGD-10 & $63649$ \Tstrut \\
         \textbf{Blacksmith} & $14137$ \Tstrut \\
    \end{tabular}
\end{table}

\section{Conclusion}
In this paper, we proposed a novel adversarial training algorithm, dubbed Blacksmith, which utilizes both single-step and multi-step ideas in a randomized manner. Blacksmith's creation was triggered by the observation that the fast adversarial training of Vision Transformers yields poorer results than their convolutional counter-parts. By presenting the Hammering and Forging schemes, insight into what Blacksmith does to improve this accuracy, as well as why, was provided. Unlike traditional single-step methods, which suffer from a much lower adversarial accuracy compared to PGD-2 when training Vision Transformers, Blacksmith achieved performance on-par with PGD-2, even overtaking it in every situation when tuned correctly.

Continuing the same line of thought as this paper, aiming to employ more PGD steps in fast methods to achieve the same level of adversarial accuracy could prove to be an interesting future research direction to explore. Moreover, studying the application of Blacksmith or similar methods to convolutional neural networks may be a fruitful endeavor.

\appendix

\bibliography{aaai24}

\appendix 

\section{Appendix}
\subsection{Results of cyclic and multi-step schedulers}\label{sec:a}

Tables~\ref{tab:CIFAR10_results_alt} and \ref{tab:CIFAR100_results_alt} present the clean and adversarial accuracies of training ViT-B with their alternative learning rate schedulers, as the perturbation size $\epsilon$ varies. The measurements provided are taken from a series of three separate experiments using different random seeds. These tables show that using cyclic learning-rate schedulers for FGSM and its variants, and using the multi-step learning-rate scheduler for PGD-2 yield sub-optimal adversarial results. I.e., the training has a higher chance of encountering the CO problem, while also suffering from a lower adversarial accuracy. Therefore, the experimental setup chosen for comparison in the paper serves as a better benchmark for evaluating Blacksmith.

\begin{table*}[h]
    \centering
    \renewcommand{\arraystretch}{1.2}
    \caption{Clean and adversarial evaluation results on the CIFAR10 dataset. Here, PGD-2 uses the multi-step scheduler and the other single-step methods use the cyclic scheduler.}
    \label{tab:CIFAR10_results_alt}
    \begin{tabular}{c|c||@{\hskip 0.125cm}c@{\hskip 0.25cm}c@{\hskip 0.25cm}c@{\hskip 0.25cm}c@{\hskip 0.25cm}c@{\hskip 0.125cm}}
        Method & Evaluation Mode & $\epsilon = 8$ & $\epsilon = 10$ & $\epsilon = 12$ & $\epsilon = 14$ & $\epsilon = 16$ \\
        \hline \\[-0.41cm] \hline
        \multirow{2}{*}{FGSM} & Clean (\%) & $51.67 \pm 7.49$ & $30.03 \pm 4.6$ & $34.59 \pm 4.51$ & $46.9 \pm 14.75$ & $31.49 \pm 4.61$ \Tstrut \\
         & Adversarial (\%) & $9.47 \pm 13.39$ & $0.0 \pm 0.0$ & $0.0 \pm 0.0$ & $8.0 \pm 11.32$ & $0.0 \pm 0.0$ \Tstrut \\
         \hline
         \multirow{2}{*}{RS-FGSM} & Clean (\%) & $38.43 \pm 6.21$ & $42.8 \pm 18.52$ & $31.33 \pm 3.24$ & $40.39 \pm 14.31$ & $24.09 \pm 3.43$ \Tstrut \\
         & Adversarial (\%) & $0.0 \pm 0.0$ & $10.37 \pm 14.66$ & $0.0 \pm 0.0$ & $7.86 \pm 11.11$ & $0.0 \pm 0.0$ \Tstrut \\
         \hline
         \multirow{2}{*}{N-FGSM} & Clean (\%) & $67.78 \pm 22.54$ & $35.02 \pm 14.73$ & $18.57 \pm 1.99$ & $17.08 \pm 5.58$ & $19.63 \pm 10.29$ \Tstrut \\
         & Adversarial (\%) & $31.49 \pm 22.27$ & $6.8 \pm 9.61$ & $0.0 \pm 0.0$ & $0.0 \pm 0.0$ & $4.34 \pm 6.14$ \Tstrut \\
         \hline
         \multirow{2}{*}{PGD-2} & Clean (\%) & $88.03 \pm 0.4$ & $85.47 \pm 0.24$ & $83.8 \pm 0.29$ & $82.25 \pm 0.05$ & $79.27 \pm 0.42$ \Tstrut \\
         & Adversarial (\%) & $41.7 \pm 0.33$ & $32.13 \pm 0.31$ & $25.57 \pm 0.12$ & $21.63 \pm 0.31$ & $20.23 \pm 0.17$  \Tstrut \\
    \end{tabular}
\end{table*}

\begin{table*}[h]
    \centering
    \renewcommand{\arraystretch}{1.2}
    \caption{Clean and adversarial evaluation results on the CIFAR100 dataset. Here, PGD-2 uses the multi-step scheduler and the other single-step methods use the cyclic scheduler.}
    \label{tab:CIFAR100_results_alt}
    \begin{tabular}{c|c||@{\hskip 0.125cm}c@{\hskip 0.25cm}c@{\hskip 0.25cm}c@{\hskip 0.25cm}c@{\hskip 0.25cm}c@{\hskip 0.125cm}}
        Method & Evaluation Mode & $\epsilon = 8$ & $\epsilon = 10$ & $\epsilon = 12$ & $\epsilon = 14$ & $\epsilon = 16$ \\
        \hline \\[-0.41cm] \hline
        \multirow{2}{*}{FGSM} & Clean (\%) & $50.83 \pm 11.21$ & $52.07 \pm 10.93$ & $43.35 \pm 5.39$ & $36.85 \pm 4.72$ & $27.1 \pm 4.91$ \Tstrut \\
         & Adversarial (\%) & $0.13 \pm 0.19$ & $6.43 \pm 9.1$ & $0.0 \pm 0.0$ & $0.0 \pm 0.0$ & $0.0 \pm 0.0$ \Tstrut \\
         \hline
         \multirow{2}{*}{RS-FGSM} & Clean (\%) & $50.08 \pm 3.7$ & $36.4 \pm 5.65$ & $35.03 \pm 3.55$ & $39.1 \pm 4.69$ & $31.54 \pm 8.8$ \Tstrut \\
         & Adversarial (\%) & $0.03 \pm 0.05$ & $0.0 \pm 0.0$ & $0.0 \pm 0.0$ & $6.56 \pm 9.28$ & $0.0 \pm 0.0$ \Tstrut \\
         \hline
         \multirow{2}{*}{N-FGSM} & Clean (\%) & $50.37 \pm 13.09$ & $50.49 \pm 11.52$ & $49.42 \pm 14.99$ & $40.08 \pm 10.02$ & $23.59 \pm 4.42$ \Tstrut \\
         & Adversarial (\%) & $16.85 \pm 16.32$ & $12.11 \pm 17.13$ & $25.25 \pm 6.21$ & $22.39 \pm 4.41$ & $10.03 \pm 7.13$ \Tstrut \\
         \hline
         \multirow{2}{*}{PGD-2} & Clean (\%) & $64.83 \pm 0.05$ & $59.43 \pm 2.57$ & $57.97 \pm 0.12$ & $55.47 \pm 0.12$ & $53.47 \pm 0.26$ \Tstrut \\
         & Adversarial (\%) & $22.1 \pm 0.08$ & $16.23 \pm 1.01$ & $13.4 \pm 0.08$ & $10.73 \pm 0.05$ & $8.93 \pm 0.05$  \Tstrut \\
    \end{tabular}
\end{table*}

\subsection{Forge-Rate}\label{sec:b}
Table~\ref{tab:forge_rate_tuning} showcases the experiments conducted to fine-tune the Forge-Rate $\lambda$ for Blacksmith. The measurements indicate that increasing $\lambda$ results in lower clean accuracy and higher adversarial accuracy, signifying a trade-off. Since increasing $\lambda$ beyond $0.66$ proved to worsen the model's performance, and since both high clean and adversarial accuracies are desired, we use $\lambda=0.66$ for training Blacksmith during the first $\frac{N}{2}$ epochs, and using $\lambda=0.33$ for the last $\frac{N}{2}$ epochs. Here, $N$ is the total training epoch count.

\begin{table*}[h]
    \centering
    \renewcommand{\arraystretch}{1.2}
    \caption{Clean and adversarial evaluation results on the CIFAR10 and CIFAR100 datasets with $\epsilon = 16/255.0$ for different Forge-Rates ($\lambda$).}
    \label{tab:forge_rate_tuning}
    \begin{tabular}{c|c|@{\hskip 0.125cm}c@{\hskip 0.25cm}c@{\hskip 0.25cm}c@{\hskip 0.25cm}c@{\hskip 0.25cm}c@{\hskip 0.125cm}}
         Dataset & Evaluation Mode & $\lambda = 0.33$ & $\lambda = 0.5$ & $\lambda = 0.66$ & our method \\
        \hline \\[-0.41cm] \hline
        \multirow{2}{*}{CIFAR10} & Clean (\%) & $73.97 $ & $72.18$ & $71.89$ & $72.49$ \Tstrut \\
         & Adversarial (\%) & $21.35$ & $22.0$ & $22.13$ & $22.04$\Tstrut \\
         \hline
        \multirow{2}{*}{CIFAR100} & Clean (\%) & $55.2 $ & $53.1$ & $50.8$  &$52.34$\Tstrut \\
         & Adversarial (\%) & $6.66$ & $11.09$ & $10.88$ &$11.21$\Tstrut \\
    \end{tabular}
\end{table*}

\subsection{PGD-2 step-size}\label{sec:c}

Table~\ref{tab:PGD_alpha_choice} showcases experimental results of using different values of step-size $\alpha$ in PGD-2, while the perturbation size varies. The measurements provided are taken from a series of three separate experiments using different random seeds. These results show that using $\alpha=\frac{\epsilon}{2}$ instead of $\alpha=\epsilon$ yields higher clean and adversarial accuracies, as well as lower standard deviations in all scenarios. Therefore, using $\alpha=\frac{\epsilon}{2}$ for PGD-2 in our main experiments makes for a stronger baseline for evaluating Blacksmith on.

\begin{table*}[h]
    \centering
    \renewcommand{\arraystretch}{1.2}
    \caption{Clean and adversarial evaluation results of PGD-2 with different choices of step-size $\alpha$ on the CIFAR10 and CIFAR100 datasets, while the perturbation size $\epsilon$ varies. The better version of PGD-2 is shown in bold-face font.}
    \label{tab:PGD_alpha_choice}
    \begin{tabular}{c|c|c|@{\hskip 0.125cm}c@{\hskip 0.25cm}c@{\hskip 0.25cm}c@{\hskip 0.25cm}c@{\hskip 0.25cm}c@{\hskip 0.125cm}}
        Dataset & Method & Evaluation Mode & $\epsilon = 8$ & $\epsilon = 10$ & $\epsilon = 12$ \\
        \hline \\[-0.41cm] \hline
         \multirow{4}{*}{CIFAR10}&\multirow{2}{*}{\textbf{PGD-2 ($\alpha = \frac{\epsilon}{2}$)}} & Clean (\%) & $89.99 \pm 0.08$ & $87.8 \pm 0.05$ & $84.31 \pm 0.13$\Tstrut \\
         & &Adversarial (\%) & $43.51 \pm 0.13$ & $35.9 \pm 0.15$ & $30.47 \pm 0.15$ \Tstrut \\
         \cline{2-6} &
        \multirow{2}{*}{PGD-2 ($\alpha = \epsilon$)} & Clean (\%) & $86.33 \pm 0.32$ & $84.4 \pm 0.26$ & $82.73 \pm 0.21$\Tstrut \\
         &  &Adversarial (\%) & $42.47 \pm 0.29$ & $33.3 \pm 0.3$ & $27.93 \pm 0.76$ \Tstrut \\
         \hline
         
         \multirow{4}{*}{CIFAR100}&\multirow{2}{*}{\textbf{PGD-2  ($\alpha = \frac{\epsilon}{2}$)}} & Clean (\%) & $68.68 \pm 0.16$ & $65.54 \pm 0.09$ & $62.71 \pm 0.12$ &\Tstrut \\
         &  & Adversarial (\%) & $23.39 \pm 0.14$ & $18.73 \pm 0.05$ & $15.5 \pm 0.11$ &  \Tstrut \\
         \cline{2-6} &
        \multirow{2}{*}{PGD-2 ($\alpha = \epsilon$)} & Clean (\%) & $62.87 \pm 0.35$ & $58.87 \pm 0.12$ & $55.97 \pm 0.6$\Tstrut \\
         &  &Adversarial (\%) & $22.67 \pm 0.12$ & $17.47 \pm 0.25$ & $13.9 \pm 0.17$ \Tstrut \\
    \end{tabular}
\end{table*}

\subsection{ViT-Small}\label{sec:d}

Table~\ref{tab:Vit_small_results} compares the clean and adversarial accuracies of Blacksmith and Blacksmith (RS) with N-FGSM and PGD-2 when training the ViT-Small model. Here, $\alpha=\frac{\epsilon}{2}$ is used for PGD-2 to make for a better baseline. Although N-FGSM achieved the best adversarial results in this setting, it exhibits the worst clean performance among the four methods, probably due to the larger applied noise. All the while, Blacksmith consistently puts forth PGD-2 level clean and adversarial accuracies, keeping its performance promises.  Moreover, Blacksmith (RS) is competitive with N-FGSM in terms of adversarial accuracy in this setting. That being said, we believe that as the ViT model's size increases, Blacksmith outshines N-FGSM more and more, while keeping up with PGD-2.

\begin{table*}[h]
    \centering
    \renewcommand{\arraystretch}{1.2}
    \caption{Clean and adversarial evaluation results on the CIFAR10 and CIFAR100 datasets, while varying the perturbation size ($\epsilon$) and training the ViT-Small model.}
    \label{tab:Vit_small_results}
    \begin{tabular}{c|c|c|@{\hskip 0.125cm}c@{\hskip 0.25cm}c@{\hskip 0.25cm}c@{\hskip 0.25cm}c@{\hskip 0.25cm}c@{\hskip 0.125cm}}
        Dataset & Method & Evaluation Mode & $\epsilon = 8$ & $\epsilon = 16$\\
        \hline \\[-0.41cm] \hline
         \multirow{8}{*}{CIFAR10}&\multirow{2}{*}{PGD-2} & Clean (\%) & $86.21 \pm 0.14$ & $69.9\pm 0.09$\Tstrut \\
         & &Adversarial (\%) & $41.74 \pm 0.13$ & $23.61\pm 0.18$ \Tstrut \\
         \cline{2-5} &
        \multirow{2}{*}{N-FGSM} & Clean (\%) & $80.96\pm 0.54$ & $49.63\pm 1.04$\Tstrut \\
         &  &Adversarial (\%) & $45.76\pm 0.42$ & $25.31 \pm 0.56$ \Tstrut \\
         \cline{2-5} 
         &\multirow{2}{*}{Blacksmith} & Clean (\%) & $85.08\pm 0.09$ & $66.76\pm 0.7$ \Tstrut \\
         &  &Adversarial (\%) & $41.88\pm 0.41$ & $22.18\pm 0.37$ \Tstrut \\
         \cline{2-5}
        &\multirow{2}{*}{Blacksmith (RS)} & Clean (\%) & $83.18\pm 0.22$ & $60.56\pm 0.89$ \Tstrut \\
         &  &Adversarial (\%) & $44.13\pm 0.31$ & $24.63\pm 0.51$ \Tstrut \\
         \hline
         
         \multirow{8}{*}{CIFAR100}&\multirow{2}{*}{PGD-2} & Clean (\%) & $62.3\pm 0.13$ & $50.74\pm 0.32$\Tstrut \\
         & &Adversarial (\%) & $19.9\pm 0.07$ & $11.35\pm 0.15$ \Tstrut \\
         \cline{2-5} &
        \multirow{2}{*}{N-FGSM} & Clean (\%) & $57.48\pm 0.44$ & $39.11\pm 0.78$\Tstrut \\
         &  &Adversarial (\%) & $23.85 \pm 0.35$ & $13.58 \pm 0.4$ \Tstrut \\
         \cline{2-5} 
         &\multirow{2}{*}{Blacksmith} & Clean (\%) & $60.64\pm 0.14$ & $47.07\pm 0.31$ \Tstrut \\
         &  &Adversarial (\%) & $21.98\pm 0.34$ & $11.34\pm 0.11$ \Tstrut \\
         \cline{2-5}
        &\multirow{2}{*}{Blacksmith (RS)} & Clean (\%) & $58.92\pm 0.33$ & $43.04\pm 0.93$ \Tstrut \\
         &  &Adversarial (\%) & $23.33\pm 0.17$ & $12.52\pm 0.13$ \Tstrut \\
    \end{tabular}
\end{table*}

\end{document}